# How to Shift Bias:
# Lessons from the Baldwin Effect

Peter Turney
Institute for Information Technology
National Research Council Canada
Ottawa, Ontario, Canada
K1A 0R6

613-993-8564
peter@ai.iit.nrc.ca

## Abstract

An inductive learning algorithm takes a set of data as input and generates a hypothesis as output. A set of data is typically consistent with an infinite number of hypotheses; therefore, there must be factors other than the data that determine the output of the learning algorithm. In machine learning, these other factors are called the *bias* of the learner. Classical learning algorithms have a fixed bias, implicit in their design. Recently developed learning algorithms dynamically adjust their bias as they search for a hypothesis. Algorithms that *shift bias* in this manner are not as well understood as classical algorithms. In this paper, we show that the Baldwin effect has implications for the design and analysis of bias shifting algorithms. The Baldwin effect was proposed in 1896, to explain how phenomena that might appear to require Lamarckian evolution (inheritance of acquired characteristics) can arise from purely Darwinian evolution. Hinton and Nowlan presented a computational model of the Baldwin effect in 1987. We explore a variation on their model, which we constructed explicitly to illustrate the lessons that the Baldwin effect has for research in bias shifting algorithms. The main lesson is that it appears that a good strategy for shift of bias in a learning algorithm is to begin with a weak bias and gradually shift to a strong bias.

# 1. Introduction

The most common learning paradigm in machine learning research is known as *supervised learning from examples*. In this paradigm, a learning algorithm is given a set of training *examples* (i.e., data, instances, cases, feature vectors) that have been classified by a *teacher* (hence *supervised* learning) into a finite number of distinct classes. The examples are usually represented as vectors in a multidimensional space, called *feature space*. The task of the learning algorithm is to induce a concept (i.e., a hypothesis, a theory, a total function) that maps from feature space into the set of classes. The concept should be consistent with the training data, but usually there is some noise in the data, so the induced concept is not required to be perfectly accurate for the training data. The quality of the induced concept is commonly measured by its accuracy on testing data, sampled from the same probability distribution as the training data.

In general, the training data do not determine a unique concept. Frequently there are an infinite number of concepts that are consistent with the data. Any factors other than the training data that determine the concept selected by the learning algorithm are called the *bias* of the algorithm (Utgoff and Mitchell, 1982; Utgoff, 1986; Rendell, 1986; Haussler, 1988; Gordon and desJardins, 1995). For example, it is common to build learning algorithms that have a bias towards simpler concepts, even when simpler concepts have lower accuracy on the training data than more complex concepts.

Bias can vary in strength; some learners have strong bias and others have weak bias. Although the word *bias* has negative connotations in general usage, strong bias has some benefits for a learning algorithm. Strong bias can reduce the size of the learner's search in concept space, which reduces the amount of computation required (Utgoff and Mitchell, 1982; Utgoff, 1986; Rendell, 1986; Haussler, 1988). Strong bias also decreases the sensitivity of the learner to noise in the data (Geman *et al.*, 1992). The cost of strong bias is that it may cause the learner to inaccurately characterize a concept, hence resulting in poor generalization.

Classical machine learning algorithms have a fixed bias, which is implicit in their design. More recent work in machine learning has been concerned with algorithms that *shift bias* dynamically, as they search for a concept (Utgoff and Mitchell, 1982; Utgoff, 1986; Tcheng *et al.,* 1989; Schaffer, 1993; Gordon and desJardins, 1995; Bala *et al.*, 1995; Provost and Buchanan, 1995; Turney, 1995). Bias shifting algorithms are not as well understood as classical algorithms. The goal of this paper is to increase our understanding of shift of bias. It is our claim that the Baldwin effect is relevant for research in the design and analysis of bias shifting algorithms.

The Baldwin effect is concerned with the evolution of populations of individuals that learn during their lifetime. The Baldwin effect was first studied by evolutionary theorists (Baldwin, 1896; Morgan, 1896; Osborn, 1896; Waddington, 1942; Maynard Smith, 1987; Wcislo, 1989; Anderson, 1995), but since the foundational paper by Hinton and Nowlan (1987), it has become an area of active research by computer scientists (Ackley and Littman, 1991; Harvey, 1993; Whitley and Gruau, 1993; Whitley *et al.,* 1994; Nolfi *et al.,* 1994; Belew and Mitchell, 1996). There are two aspects to the Baldwin effect. First, learning smooths the fitness landscape, which can accelerate evolution. Second, learning has a cost, so there is evolutionary pressure to find instinctive replacements for learned behaviors, in stable environments. When a population evolves a new behavior, the Baldwin effect predicts that the evolution will pass through two phases. In the early phase, there will be a selective pressure in favor of learning, due to the first aspect of the Baldwin effect. In the later phase, there will be a selective pressure in favor of instinct, due to the second aspect of the Baldwin effect.

One of the main points we argue in this paper is that strong bias is analogous to instinct and weak bias is analogous to learning. Therefore the Baldwin effect predicts that, under certain conditions (some of the conditions are discussed later in the paper), a bias shifting algorithm will begin with weak bias, but gradually move to strong bias. We argue here that the Baldwin effect is not merely a *description* or prediction of the behavior of a bias shifting algorithm; it is also a *prescription*. The Baldwin effect does not merely predict the trajectory of a bias shifting algorithm as it searches through bias space. We show that the predicted trajectory is superior to some alternative trajectories. This is an interesting lesson for the design of bias shifting algorithms.

In the experiments we present here, the search in bias space is done by an evolutionary algorithm. However, we believe that the lessons generalize to other methods for search in bias space. Although we have not tested this conjecture, the phenomena we investigate here do not seem to depend on any special properties of evolutionary computation. It appears that the Baldwin effect is not specifically concerned with evolution. In its most general sense, the Baldwin effect is concerned with simultaneous search in two distinct but related spaces. An open problem is to discover the most general conditions for the manifestation of the Baldwin effect. Evolution (as opposed to search in a more general sense) does not appear to be a necessary condition.

In the next section, we discuss the Baldwin effect in more detail. We review the experiment of Hinton and Nowlan (1987) and the refinement of this experiment by Harvey (1993). In Section 3, we examine the concepts of bias and bias shifting algorithms. Section 4 presents a simple model of the evolution of bias, along the lines of the experiments of Hinton and Nowlan (1987) and Harvey (1993). The model is designed to draw out some of the implications of Hinton and Nowlan (1987) for research in bias shifting algorithms. Experiments with this model (Section 5) show that the Baldwin effect has interesting lessons for bias shifting. We discuss the implications of this work in Section 6 and we conclude in Section 7.

## 2. The Baldwin Effect

This section begins with a discussion of the two aspects of the Baldwin effect. It then reviews the computational model of Hinton and Nowlan (1987). Their work clearly illustrates the first aspect of the Baldwin effect, but the second aspect is not as clearly apparent. Harvey (1993) analyzes the model in detail with respect to the second aspect of the Baldwin effect.

### 2.1 Two Aspects

Baldwin (1896) proposed 100 years ago that learning could combine synergistically with evolution. Similar ideas were proposed by Morgan (1896) and Osborn (1896). When Baldwin, Morgan, and Osborn were writing, there was active debate between Darwinian and Lamarckian theories of evolution. The Lamarckians believed in the inheritance of acquired characteristics. They argued that paleontology supported their belief, because the fossil record shows rapid change that cannot be explained by Darwinian evolution (this argument has come back to us in the more recent controversy over punctuated equilibria). Baldwin, Morgan, and Osborn did not have the current theory of the molecular mechanisms of genetics, which makes the inheritance of acquired characteristics appear highly unlikely. Instead, they replied to Lamarckian criticism by showing how learning could interact with Darwinian evolution in a way that could account for phenomena that appear to require Lamarckian evolution.

Baldwin (1896) was particularly concerned with explaining how complex instinctive behaviors could evolve. Darwinian evolution requires a complex instinct to evolve in small, incremental steps. But what survival advantage could come from a partial instinct? Baldwin's proposal has two aspects. (1) Learning can enable an organism to repair deficiencies in a partial instinct. Thus a partial instinct can still enhance fitness, if it is combined with learning. (2) However, learning can be costly to organisms. Learning takes time and energy and usually involves making mistakes. Thus, over many generations, learned behaviors may be replaced by instinctive behaviors.

The Baldwin effect appears similar to Lamarckian evolution. Behaviors that are learned by an individual (acquired characteristics) become instinctive (inherited characteristics) over a period of many generations. However, unlike Lamarckian evolution, there is no need to postulate a mechanism for directly altering the genotype (e.g., the DNA) of an organism, based on the experience of the phenotype (e.g., what the organism has learned).

The first aspect of the Baldwin effect is that learning can accelerate evolution by permitting partially successful genotypes to reproduce. In modern terminology, learning smooths the fitness landscape (Maynard Smith, 1987; Hinton and Nowlan, 1987; Whitley and Gruau, 1993; Whitley *et al.*, 1994). We may view learning as local search on the fitness landscape, in the neighborhood around a given genotype. The fitness of the phenotype is determined by the result of this local search, rather than merely by the genotype. This aspect of the Baldwin effect appeals to computer scientists, because it can increase the effectiveness of problem solving by evolutionary computation (Ackley and Littman, 1991; Whitley and Gruau, 1993; Whitley *et al.*, 1994; Nolfi *et al.,* 1994; Bala *et al.*, 1995; Belew and Mitchell, 1996).

The second aspect of the Baldwin effect is that learning is costly, and there are situations where instincts may be better than learning. In a dynamic situation, learning can adapt to phenomena that change too rapidly for evolutionary adaptation. Instincts will not evolve in these situations. However, if a situation is relatively stable, then, given enough time, we should expect evolution to eventually generate an instinctive replacement for the learned behavior. This second aspect of the Baldwin effect has received less attention from computer scientists than the first aspect (Belew and Mitchell, 1996; Hinton and Nowlan, 1987; Anderson, 1995), but it has interesting consequences for problem solving by evolutionary computation.

## 2.2  Hinton and Nowlan's Experiment

Hinton and Nowlan (1987) constructed the first computational model of the Baldwin effect. They deliberately made a very simple model, to show more clearly the mechanisms of the effect. The model has a population of 1000 individuals with genotypes composed of 20 genes. The genes have the values **0**, **1**, or **?**. A genotype is interpreted as specifying the setting of 20 switches. If a gene has the value **0**, then the individual must open the corresponding switch. If a gene has the value **1**, then the individual must close the corresponding switch. If a gene has the value **?**, then the individual is permitted to experiment with setting the corresponding switch open or closed. In the first generation, the genes are randomly set to **0**, **1**, or **?**, with probabilities 0.25, 0.25, and 0.5, respectively. The fitness $F$ of an individual is 1, unless the individual happens to close all of the switches. The individual is allowed 1000 random guesses at the correct setting (closed) of the switches that have genes with the value **?**. If the individual closes all of the switches after $i$ guesses, then its fitness is calculated by the formula $F = 1 + 19(1000 - i)/1000$. If a genotype is all **1**s, then $i = 0$ and $F = 20$, which is the maximum fitness. If an individual has the misfortune of having a **0** gene, its fitness must be 1, the minimum fitness. Similarly, if an individual does not happen to close all of the switches within 1000 guesses, then $i = 1000$ and $F = 1$. The expected fitness of an individual rises as the number of **1** genes rises, assuming the individual has no **0** genes.

Hinton and Nowlan (1987) allowed the population to evolve for 50 generations. They observed that **0** genes were rapidly eliminated from the population and the frequency of **1**s rapidly increased. In a plot of gene frequency as a function of the generation number (their Figure 2, not reproduced here), it appears that, after 50 generations, the frequency of **?**s is stable at 45% and the frequency of **1**s is stable at 55%. With twenty genes, an average individual in the final generation would have eleven **1** and nine **?** genes. As Harvey (1993) shows, the expected fitness of such an individual is roughly $F = 11.6$.

Hinton and Nowlan (1987) argue for the value of the Baldwin effect by considering what would happen in this model if there were no learning. We may interpret the number of **?** genes in a genotype as a measure of the degree of learning in the individual. An individual with no learning (only **0**s and **1**s; no **?**s) would have a fitness of 1, unless all of its genes were **1**, in which case it would have a fitness of 20. If **1**s and **0**s are generated randomly with equal probability, the chance of twenty **1**s is:

$$\frac{1}{2^{20}} = \frac{1}{1,048,576} \tag{1}$$

A population of 1000 individuals evolving over 50 generations results in only 50,000 trials, so it is highly unlikely that one of the individuals will have only **1**s in its genotype when **?** genes are not allowed. Thus, without learning, the expected fitness in the final generation is about $F = 1$. (A more precise argument by Harvey (1993) shows that $F = 1.009$.)

This experiment illustrates the first aspect of the Baldwin effect. With learning, Hinton and Nowlan (1987) observe an average fitness of about $F = 11.6$ after 50 generations. Without learning, the expected fitness is about $F = 1$. It is clear that learning aids the evolutionary process in this case.

## 2.3  Harvey's Experiment

Harvey (1993) addressed "the puzzle of the persistent question marks". That is, in Hinton and Nowlan's (1987) experiment, why does it appear that the frequency of **?**s is stable at 45% and the frequency of **1**s is stable at 55%? We expect the frequency of **?**s to eventually approach 0% and the frequency of **1**s to approach 100%. Instead, the population appears to have settled in a suboptimal state with an average fitness of $F = 11.6$, which is far from the $F = 20$ that is possible.

Part of the answer to this puzzle is that Hinton and Nowlan (1987) only ran their model for 50 generations. In Figure 1 (from Harvey (1993)), we see what happens when the model runs for 500 generations: the frequency of **?**s continues to fall and the frequency of **1**s continues to rise. However, Harvey (1993) found that the frequency of **?**s never reaches 0%. The reason is *genetic drift*, due to random mutation in the population. Mutation exerts a constant pressure that maintains a certain frequency of **?**s in the population. The population eventually achieves an equilibrium state where the pressure of genetic drift balances with the selection pressure that favors **1**s over **?**s.

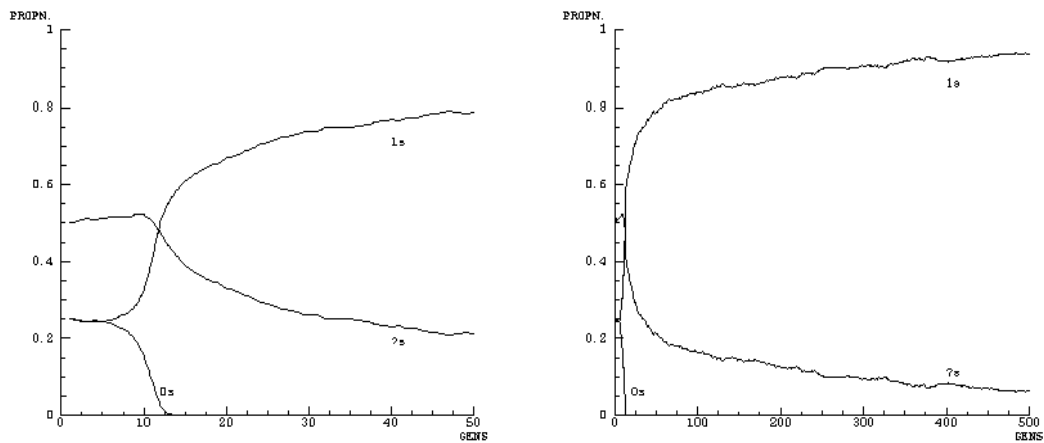

Figure 1. "The proportions of incorrect, correct, and undecided (adaptive) alleles (**0**s, **1**s, **?**s) in the whole population, against generations. On the left, the first 50 generations of a run, and on the right the same continued for 500 generations." From Harvey (1993).

In Figure 1, we see an initial rise in the frequency of **?**s, followed by a decline. This is exactly what we should expect from the Baldwin effect. The frequency of **?**s is a measure of the degree of learning. Initially there is a selective pressure for learning, which we see in the rise in the frequency of **?**s. This is the first aspect of the Baldwin effect. Later there is a selective pressure for instinct, which we see in the fall in the frequency of **?**s. This is the second aspect of the Baldwin effect.

# 3. Bias

We begin this section with a definition of bias and then discuss algorithms that shift bias dynamically. We examine the costs and benefits of strong bias and we conclude with some examples of bias in neural networks.

## 3.1  Definition of Bias

Excluding the input data, all the factors that influence the selection of one particular concept constitute the *bias* of a learning algorithm (Utgoff and Mitchell, 1982; Utgoff, 1986; Haussler, 1988; Gordon and desJardins, 1995). Bias includes such factors as the language in which the learner expresses its concepts, the algorithm that the learner uses to search through the space of concepts, and the criteria for deciding whether a concept is compatible with the training data.

Utgoff (1986) describes two central properties of bias: *correctness* and *strength*. A *correct* bias is a bias that allows the concept learner to select the target concept. The correctness of a bias is usually measured by the performance of the learned concept on an independent set of testing data. A *strong* bias is a bias that focuses the concept learner on a relatively small number of concepts.

Rendell (1986) distinguishes *exclusive* and *preferential* bias. A learner with an exclusive bias refuses to consider certain concepts. For example, an *exclusive* bias may be implemented by restricting the language in which the learner represents its concepts. If the language used by one learner is a subset of the language used by a second learner, then (all else equal) we may say that the first learner has a stronger bias than the second learner. A learner with a *preferential* bias prefers some concepts over other concepts. However, given sufficient evidence (in the training data), the learner may eventually select the less favored concept. We can measure the strength of preferential bias by the amount of evidence (the number of training examples) required to overcome the bias.

Statistical bias is defined somewhat differently from bias in machine learning. We discuss this in detail in the Appendix.

### 3.2  Shift of Bias

A growing body of research in machine learning is concerned with algorithms that shift bias as they acquire more experience (Utgoff and Mitchell, 1982; Utgoff, 1986; Tcheng *et al.,* 1989; Schaffer, 1993; Gordon and desJardins, 1995; Bala *et al.*, 1995; Provost and Buchanan, 1995; Turney, 1995). Shift of bias involves two levels of search: search through concept space and search through bias space. Although bias shift is often discussed in the context of incremental learners (Utgoff, 1986), it applies equally well to batch learners (Tcheng *et al.,* 1989; Schaffer, 1993; Gordon and desJardins, 1995; Bala *et al.*, 1995; Provost and Buchanan, 1995; Turney, 1995). Typically an incremental learner will shift its bias as new training cases appear, while a batch learner will shift its bias as it makes consecutive passes through the training data.

Algorithms that shift their bias do not avoid the conservation theorems ("no free lunch" theorems) that have recently received much attention in the machine learning community (Schaffer, 1994; Wolpert, 1992, 1994). The conservation theorems show that there is no single bias that works well under all circumstances. In essence, the idea is that, for every environment in which a given bias is suitable, there is another environment in which the bias performs poorly. However, algorithms that shift bias have been shown empirically to perform well under a wider range of real-world conditions than algorithms with fixed bias (Tcheng *et al.,* 1989; Schaffer, 1993; Gordon and desJardins, 1995; Bala *et al.*, 1995; Provost and Buchanan, 1995; Turney, 1995).

### 3.3  Costs and Benefits of Strong Bias

Strong bias has both costs and benefits. The major cost is that a bias can be incorrect. When a bias is incorrect, it is clearly better that it be weak than that it be strong. In general, it is risky for a learning algorithm to have a strong bias, unless we have high confidence in the correctness of the bias for the learning problems where we intend to apply the algorithm. We usually do not have such confidence, so it would seem that we should prefer weak bias to strong bias.

On the other hand, strong bias has benefits. It makes search in concept space more efficient and thereby accelerates learning. It reduces the sensitivity of the learner to noise in the training data and thereby decreases the variance of the induced concept. The expected prediction error of the learner (the expected error on the testing data) is determined by the error on the training data and the variance of the induced concept. Strong bias increases training set error but decreases variance. Weak bias decreases training set error but increases variance. This phenomenon is familiar to statisticians as the bias/variance dilemma (Geman *et al.*, 1992). The ideal bias is both strong (low variance; good resistance to noise) and correct (low training set error; low testing set error). Achieving both of these properties can be very difficult. (See the Appendix for more discussion of bias and variance.)

### 3.4  Bias in Neural Networks

Much of the research in hybrids of evolutionary computation and learning has been concerned with neural network models of learning (Balakrishnan and Honavar, 1995); therefore, it is worth discussing bias in the context of neural networks. In a typical hybrid of evolutionary computation and neural networks, evolutionary search is responsible for determining the architecture of the network (the connections and the size of the hidden layer) and back propagation is responsible for determining the appropriate weights for the connections in the network. Since the architecture of a network plays a central role in determining the bias of the network, these hybrids are examples of bias shifting algorithms. The evolutionary algorithm searches in bias space while back propagation searches in concept (function) space.

A major factor determining the bias of a network is the number of weights in the network. In general, the bias of a neural network is inversely proportional to the number of weights. Increasing the number of weights results in decreased bias, but increased variance (Geman *et al.*, 1992). Varying the architecture of a network is perhaps the most obvious way to alter its bias. However, there are many other ways to alter the bias of a network. Some examples are soft weight-sharing (Nowlan and Hinton, 1992), weight decay (Hinton, 1986; Weigend *et al.,* 1990), and weight elimination (Weigend *et al*., 1991; Weigend and Rumelhart, 1994). A significant body of research in neural networks addresses the problem of adjusting the bias of a network.

## 4. A Model of the Baldwin Effect and Bias Shift

We would like to build a simple computational model, along the lines of Hinton and Nowlan's (1987) model, that clearly illustrates both the Baldwin effect and shift of bias. Shift of bias is already implicit in Hinton and Nowlan's (1987) model, although it is perhaps not obvious. A genotype with many **?**s has a weak bias and a genotype with many **0**s and **1**s has a strong bias. The plot of the frequency of **?**s (Figure 1) is essentially a plot of the trajectory of the search of the population in bias space. The following variation on Hinton and Nowlan's (1987) model should help to make this more clear.

Let us consider a simple example of concept learning. Suppose the examples consist of five-dimensional Boolean vectors $\vec{x} \in \{0, 1\}^5$ that each belong to one of two classes $\{0, 1\}$. The space of concepts is the space of functions $\{0, 1\}^5 \rightarrow \{0, 1\}$ mapping vectors to classes. Each possible concept can be identified with its truth-table. A truth-table lists each of the $2^5 = 32$ possible vectors and the value of the function for each vector. If we assume a standard order for listing the 32 input vectors $\vec{x} \in \{0, 1\}^5$, then we can compactly code each function by the column in the truth-table that corresponds to the value of the function. That is, we can code each of the $2^{32} = 4{,}294{,}967{,}296$ possible functions with a 32 bit string. For example, the function that maps $\vec{x} \in \{0, 1\}^5$ to 1, for all values of $\vec{x}$, would be coded as "11111111111111111111111111111111".

Suppose that there is one particular concept (function) that is our *target* function. To facilitate comparison with Harvey (1993), we may suppose the target function is $f(\vec{x}) = 1$. We assume the standard supervised learning paradigm, with a training phase followed by a testing phase. During training, the learner is taught the class (0 or 1) of each of the 32 possible input vectors. However, there is a certain probability *p* that the learner is taught the wrong class. During testing, the learner must guess the class of each of the possible input vectors. Again, there is a probability $p$ that the tester is mistaken about the correct class for an input vector. That is, the probability $p$ is the level of noise in the class.

$$\text{target} = \langle t_1, ..., t_{32} \rangle = \vec{t} \tag{2}$$

$$\text{train} = \langle \alpha_1, ..., \alpha_{32} \rangle = \vec{\alpha} \tag{3}$$

$$\text{test} = \langle \beta_1, ..., \beta_{32} \rangle = \vec{\beta} \tag{4}$$

$$t_i, \alpha_i, \beta_i \in \{0, 1\} \tag{5}$$

We generate $\vec{\alpha}$ and $\vec{\beta}$ from $\vec{t}$ by randomly flipping bits in $\vec{t}$ with probability $p$. The probability that the class of a training example or a testing example matches the target is $1 - p$, but the probability that the class of the training example matches the class of the testing example is $1 - 2p + 2p^2$.

$$P(\alpha_i = t_i) = 1 - p \tag{6}$$

$$P(\beta_i = t_i) = 1 - p \tag{7}$$

$$P(\alpha_i = \beta_i) = 1 - 2p + 2p^2 \tag{8}$$

Either $\alpha_i = \beta_i = t_i$, with probability $(1-p)^2$, or $\alpha_i = \beta_i \neq t_i$, with probability $p^2$, which yields $(1-p)^2 + p^2 = 1 - 2p + 2p^2$. This kind of model is very common in statistics. The model is essentially that the observed class ($\vec{\alpha}$ or $\vec{\beta}$) is composed of a signal ($\vec{t}$) plus some random noise ($p$).

This simplified model avoids some interesting issues. There are only 32 cases to learn and we assume that all of the cases are present in the training data. Also, the representation of concepts by their truth-tables clearly does not scale up, since a Boolean function of $n$ arguments requires $2^n$ bits to encode. However, the model has enough complexity to display interesting behavior and to give us insight into the Baldwin effect and shift of bias. More complexity would obscure the phenomena we are investigating here.

Suppose a genotype has 64 genes, 32 that determine bias direction and 32 that determine bias strength. The bias direction genes are either 0 or 1. The bias strength genes are real values from 0 to 1 (coded with 8 bits each).

$$\text{genotype} = \langle d_1, ..., d_{32}, s_1, ..., s_{32} \rangle \quad (9)$$

$$\text{bias direction} = \langle d_1, ..., d_{32} \rangle \quad (10)$$

$$\text{bias strength} = \langle s_1, ..., s_{32} \rangle \quad (11)$$

$$d_i \in \{0, 1\} \quad (12)$$

$$0 \leq s_i \leq 1 \quad (13)$$

If the $i$-th bias strength gene has a value $s_i$, then there is a probability $s_i$ that the individual will guess $d_i$; otherwise, there is a probability $1 - s_i$ that the individual will guess $\alpha_i$.

$$\text{guess} = \langle g_1, ..., g_{32} \rangle = \vec{g} \quad (14)$$

$$g_i \in \{0, 1\} \quad (15)$$

$$P(g_i = d_i | d_i \neq \alpha_i) = s_i \quad (16)$$

$$P(g_i = \alpha_i | d_i \neq \alpha_i) = 1 - s_i \quad (17)$$

$$P(g_i = d_i = \alpha_i | d_i = \alpha_i) = 1 \quad (18)$$

If the bias is weak ($s_i = 0$), the individual guesses based on what it observed in the training data (it guesses $\alpha_i$). If the bias is strong ($s_i = 1$), the individual ignores the training data and relies on instinct (it guesses $d_i$).

This simplified model does not describe the learning mechanism. We are dealing with a level of abstraction where the exact learning mechanism is not important. Since we assume that all 32 possible cases are in the training data $\vec{\alpha}$, the individual can learn by simply storing the training data. In a more complex model, the genotype could encode the architecture of a neural network, and back propagation could be used to learn from the training data (Balakrishnan and Honavar, 1995).

If an individual relies entirely on instinct $(\forall i)[s_i = 1]$ and its instinct is correct $(\forall i)[d_i = t_i]$, then the probability that it will correctly classify all 32 input vectors in the testing phase is $(1 - p)^{32}$:

$$[(\forall i)[s_i = 1] \wedge (\forall i)[d_i = t_i]] \rightarrow [P((\forall i)[g_i = \beta_i]) = (1 - p)^{32}] \quad (19)$$

If an individual relies entirely on learning $(\forall i)[s_i = 0]$, then the probability that it will correctly classify all 32 testing vectors is $(1 - 2p + 2p^2)^{32}$:

$$[(\forall i)[s_i = 0]] \rightarrow [P((\forall i)[g_i = \beta_i]) = (1 - 2p + 2p^2)^{32}] \qquad (20)$$

For convenience, we want our fitness score to range from 0 (low fitness) to 1 (high fitness). To make the problem challenging, we require the individual to correctly guess the class of all 32 testing examples. (This is analogous to Hinton and Nowlan's (1987) requirement that the individual close all 20 switches to get a fitness score above the minimum.) We assign a fitness score of 0 when the guess does not perfectly match the testing data and a score of $(1 - p)^{-32}$ when the match is perfect.

$$\text{fitness} = F(\vec{g}) = \begin{bmatrix} (1-p)^{-32} & \text{if } (\forall i)[g_i = \beta_i] \\ 0 & \text{if } (\exists i)[g_i \neq \beta_i] \end{bmatrix} \qquad (21)$$

Table 1 shows the expected fitness for pure learning (bias strength 0) and pure correct instinct (bias strength 1 and bias direction equal to the target) for some sample noise levels. We see that the advantage of correct instinct over pure learning increases with increasing noise levels. (The trick is that the instinct must be correct $(\forall i)[d_i = t_i]$.)

Table 1. Comparison of two different strategies for achieving good fitness scores.

| Noise Level $p$ | Pure Learning $(\forall i)[s_i = 0]$ : | | | Pure Correct Instinct $(\forall i)[s_i = 1] \wedge (\forall i)[d_i = t_i]$ : | | |
|---|---|---|---|---|---|---|
| | Probability of Perfect Test Score | Fitness Score for Perfect Test Score | Expected Fitness | Probability of Perfect Test Score | Fitness Score for Perfect Test Score | Expected Fitness |
| 0.1% | 0.94 | 1.03 | 0.97 | 0.97 | 1.03 | 1.00 |
| 0.5% | 0.73 | 1.17 | 0.85 | 0.85 | 1.17 | 1.00 |
| 1% | 0.53 | 1.38 | 0.73 | 0.72 | 1.38 | 1.00 |

We measure bias correctness by the frequency with which the bias direction matches the target.

$$\text{bias correctness} = \frac{1}{32} \sum_{i=1}^{32} \begin{bmatrix} 1 \text{ if } (d_i = t_i) \\ 0 \text{ if } (d_i \neq t_i) \end{bmatrix} \quad (22)$$

We measure bias strength by the average value of $s_i$.

$$\text{bias strength} = \frac{1}{32} \sum_{i=1}^{32} s_i \quad (23)$$

The Baldwin effect predicts that, initially, when bias correctness is low, selection pressure will favor weak bias. Later, when bias correctness is improving, selection pressure will favor strong bias.

We can view the genotype in Hinton and Nowlan's (1987) model as a special case of our genotype:

$$\mathbf{0} \Leftrightarrow \langle d_i = 0, s_i = 1 \rangle \quad (24)$$

$$\mathbf{1} \Leftrightarrow \langle d_i = 1, s_i = 1 \rangle \quad (25)$$

$$\mathbf{?} \Leftrightarrow \langle d_i = 0, s_i = 0 \rangle \Leftrightarrow \langle d_i = 1, s_i = 0 \rangle \quad (26)$$

Our genotype generalizes Hinton and Nowlan's (1987) genotype by allowing $0 \leq s_i \leq 1$ instead of $s_i \in \{0, 1\}$. In Hinton and Nowlan's (1987) genotype, the only way to increase (decrease) bias strength is to change some **?**s to **1**s or **0**s (change some **1**s or **0**s to **?**s), which necessarily also changes bias direction. In our genotype, we can alter bias strength without changing bias direction, which we will do in some of the following experiments.

As we discussed in Section 3.3, the benefits of strong bias are reduced search and increased resistance to noise. Hinton and Nowlan's (1987) model dealt with the benefit of reduced search. Their fitness function ($F = 1 + 19(1000 - i)/1000$) increased as the amount of search decreased. Our model does not consider the amount of search; instead, it deals with the benefit of increased resistance to noise. This makes our model an interesting complement to Hinton and Nowlan's (1987) model. One attractive feature is that we can easily adjust the selection pressure for instinct (strong bias) by adjusting the noise level. Of course, we could add adjustable parameters to Hinton and Nowlan's (1987) fitness function (e.g., $F = 1 + 19(1000 - Ki)/1000$), but this seems somewhat *ad hoc*. (It is also perhaps worth noting that researchers in machine learning, neural networks, and statistics seem generally more interested in the bias/variance dilemma (Geman *et al.*, 1992) than in the efficiency of search in concept space. The assumption is that, in industrial applications, testing set accuracy is typically more important than computation time.)

# 5. Experiments

We begin this section with a description of our experimental design. Our first experiment shows that our model follows the trajectory predicted by the Baldwin effect. The next experiment shows that the behavior of the model is robust, when the first generation is deliberately skewed to make the task harder. The remaining experiments examine what happens when the population is forced to follow certain trajectories in bias space. These experiments show that the trajectory recommended by the Baldwin effect is superior to the other trajectories that we examine.

## 5.1 Experimental Design

In the following experiments, we use Grefenstette's (1983; 1986) GENESIS genetic algorithm software. We follow Hinton and Nowlan (1987) and Harvey (1993) in using a population size of 1000. To ensure that the population has evolved to an equilibrium state (where genetic drift and selective pressure are in balance), we run the model for 10,000 generations (compared to 50 for Hinton and Nowlan (1987) and 500 for Harvey (1993)). We use a crossover rate of 0.6 and a mutation rate of 0.001 (these are the default settings in GENESIS). Each experiment is repeated with three different levels of noise in the training data (that is, three values for *p*), 0.1%, 0.5%, and 1% (see Table 1 above).

For each experiment, we plot the population average for bias correctness, bias strength, and fitness as a function of the generation number. We use a logarithmic scale for the generation number, because the first aspect of the Baldwin effect (selection for learning) tends to take place quite rapidly, in the early generations, while the second aspect (selection for instinct) tends to take place much more slowly, in the later generations. This is clear in Figure 1, where Harvey (1993) was compelled to produce two plots with two different time scales. A logarithmic time scale lets us see both aspects of the Baldwin effect in one plot.

In the following experiments, we are interested in qualitative behavior more than quantitative behavior. We have found that the qualitative behavior is extremely robust across repeated runs. Therefore (following Hinton and Nowlan (1987) and Harvey (1993)) instead of averaging several runs together, we plot one typical run for each experiment. This yields a more accurate impression of the level of noise in the plots than we would get from averaging runs.

## 5.2  Baseline Experiment

In this experiment, we begin with a random initial population and allow the population to evolve for 10,000 generations. Our hypotheses were (1) bias strength will initially fall and then later rise, (2) bias strength will eventually reach an equilibrium with genetic drift, (3) the equilibrium point will be higher with higher noise levels, since higher noise levels increase the selection pressure for strong bias, and (4) fitness will begin at 0 and slowly approach 1, but genetic drift will prevent it from ever reaching 1. The first prediction follows from the two aspects of the Baldwin effect. The other three predictions follow from Harvey's (1993) analysis of the role of genetic drift in Hinton and Nowlan's (1987) model. These hypotheses are supported in Figure 2 and Table 2.

Table 2. Rate of evolution and equilibrium state for Experiment 1.

|  | First generation with: |  | Number of generations difference | Average of the last 1000 generations: |  |  |
| --- | --- | --- | --- | --- | --- | --- |
| Noise Level | Fitness above 0.0 | Fitness above 0.5 |  | Fitness | Bias Strength | Bias Correctness |
| 0.1% | 5 | 62 | 57 | 0.97 | 0.54 | 1.00 |
| 0.5% | 5 | 72 | 67 | 0.94 | 0.71 | 1.00 |
| 1% | 5 | 69 | 64 | 0.92 | 0.79 | 1.00 |

We include Table 2 mainly as a baseline for comparison with the following experiments. Columns two to four show how quickly the fitness rises. The speed does not seem to be sensitive to the noise level. (The same random number seed was used in these three runs, so the first individual with a non-zero fitness score arises in the same generation.

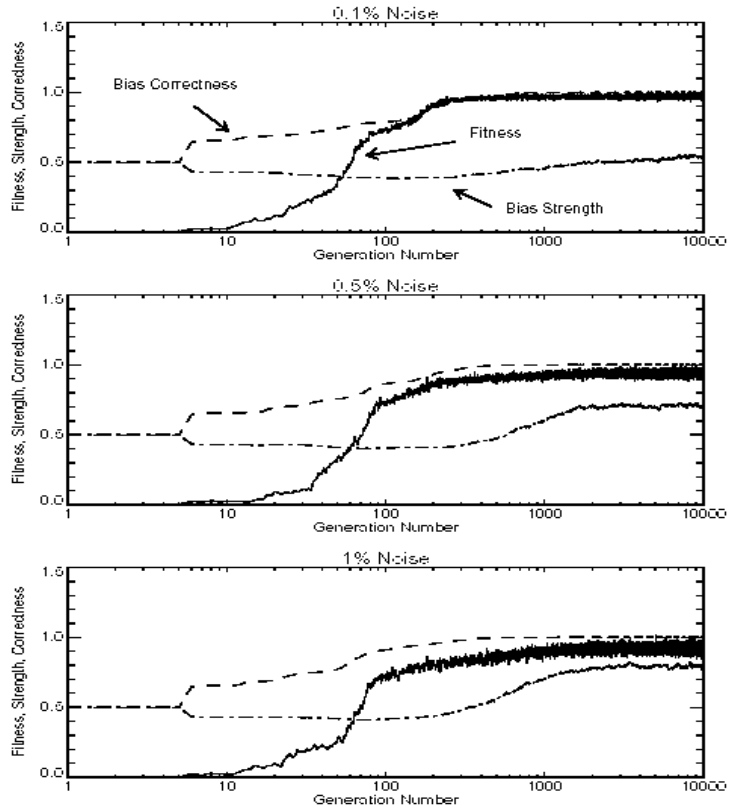

Figure 2. Experiment 1: The average fitness, bias strength, and bias correctness of a population of 1000, plotted for generations 1 to 10,000, with three different noise levels.

This helps to ensure that any differences among the three plots are due to the noise level, rather than chance. This is not required for the experiments, but it facilitates interpretation of the plots.) On the other hand, column five shows that the final fitness score is sensitive to the noise level. We might expect the final fitness to increase as the final bias strength (column six) increases, with increasing noise levels. However, the bias strength does not increase enough to counter the negative impact of noise on fitness. Still, evolved bias is performing better than pure learning (compare column five in Table 2 to column four in Table 1 — these two columns are reproduced side-by-side later, in Table 5).

## 5.3 Strong Bias in the First Generation

The following experiment is intended to test the robustness of the phenomena observed in Experiment 1. In this experiment, we deliberately skewed the first generation. Instead of using a uniform random distribution to generate the genotypes in the first generation, we used a random distribution that is skewed towards stronger bias. The bias direction genes $d_i$ were generated using equal probability for 0 and 1, but the bias strength genes $s_i$ were generated so that there was a 75% probability that $0.9 \leq s_i \leq 1.0$, a 20% probability that $0.5 \leq s_i \leq 0.9$, and a 5% probability that $0.0 \leq s_i \leq 0.5$.

Our hypotheses were (1) the population will eventually settle into (approximately) the same equilibrium state that was observed in Experiment 1, (2) the higher bias strength in the first generation will slow the creation of the first individual with a non-zero fitness, (3) but once that individual is created, there will be little difference between the two experiments, (4) during the period for which all individuals have zero fitness, genetic drift will push bias strength towards 0.5, and (5) after the discovery of the first individual with non-zero fitness, bias strength will continue to fall for a few generations, before it finally begins to rise toward the equilibrium value. These hypotheses are supported in Figure 3 and Tables 3 and 4.

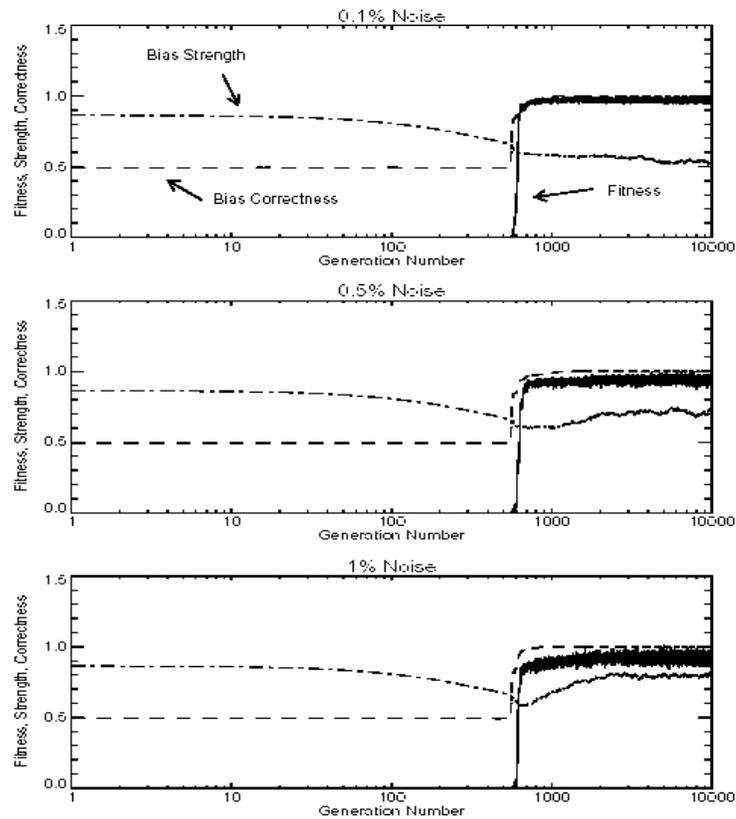

Figure 3. Experiment 2: The first generation is skewed towards stronger bias.

Table 3. Rate of evolution and equilibrium state for Experiment 2.

| Noise Level | First generation with: | | Number of generations difference | Average of the last 1000 generations: | | |
|---|---|---|---|---|---|---|
| | Fitness above 0.0 | Fitness above 0.5 | | Fitness | Bias Strength | Bias Correctness |
| 0.1% | 559 | 612 | 53 | 0.97 | 0.53 | 1.00 |
| 0.5% | 559 | 629 | 70 | 0.94 | 0.71 | 1.00 |
| 1% | 559 | 614 | 55 | 0.93 | 0.81 | 1.00 |

Table 4. Bias strength at various stages of evolution.

| Noise Level | Bias Strength: | | | |
|---|---|---|---|---|
| | Generation 1 | When Fitness is first above 0.0 | When Fitness is first above 0.5 | Average of last 1000 generations |
| 0.1% | 0.86 | 0.66 | 0.60 | 0.53 |
| 0.5% | 0.86 | 0.66 | 0.62 | 0.71 |
| 1% | 0.86 | 0.66 | 0.59 | 0.81 |

Table 3 has almost the same values in the last three columns as we observed in Experiment 1 (Table 2), which shows that the final equilibrium state is robust to large perturbations of the initial generation. Also, we see that the number of generations between the creation of the first non-zero fitness individual and a population average fitness above 0.5 is (approximately) the same in both experiments (compare column 4 in Table 3 with column 4 in Table 2).

Table 4 shows that bias strength starts at 0.86 in the first generation and slowly drifts down to 0.66. When the average bias strength reaches 0.66, the bias is weak enough for learning to compensate for incorrect bias direction, and an individual with non-zero fitness is generated. Even after this individual arises, there is selective pressure for weak bias, so bias strength continues to drop, until the population average fitness is about 0.5. At this point, bias correctness is close to 1.0, so the second aspect of the Baldwin effect begins to dominate, and there is increased selective pressure for strong bias. Bias strength rises until it reaches equilibrium with genetic drift, where the equilibrium point depends on the noise level and the mutation rate.

## 5.4 Forced Bias Strength Trajectories

So far, the experiments here do not go significantly beyond Harvey's (1993) work. Both here and in Harvey (1993), the essential point has been to show that a computational model can manifest the two aspects of the Baldwin effect. Perhaps the only important difference is that our model has been developed to draw out the implications of the Baldwin effect for bias shifting algorithms. Harvey's (1993) work has essentially the same relevance for bias shifting, although it is perhaps less readily apparent.

So far, we have only demonstrated that the Baldwin effect is a good *description* of our model; it makes accurate predictions about the behavior of the model. The Baldwin effect becomes more interesting when we consider it *prescriptively*. The Baldwin effect accurately predicts that evolution first selects for weak bias and then later selects for strong bias, but is this the *best* trajectory through bias space? What if we could force the population to follow another trajectory through bias space?

Hinton and Nowlan (1987) partially addressed this question by considering one other possible trajectory through bias space. They considered what would happen if there were no **?**s in the genotype. They observed that the chance of discovering the only genotype with a non-minimal fitness score would be 1 in $2^{20} = 1,048,576$, which is not very likely in 50 generations of a population of 1000. In our model, the analogous question is what would happen if the bias strength were frozen at 1, $(\forall i)[s_i = 1]$ ? The chance of matching the testing data (the chance of a non-zero fitness score) would be 1 in $2^{32} = 4,294,967,296$, which is not very likely in 10,000 generations of a population of 1000. But this only shows that one particular trajectory through bias space (the maximum bias strength trajectory) is inferior to a Baldwinian trajectory (start weak, end strong). There are many other trajectories that we would like to compare with a Baldwinian trajectory.

Table 5 summarizes what we know so far about the expected fitness resulting from different trajectories through bias space. Pure instinct has $1,000 \times 10,000 = 10,000,000$ chances in $2^{32} = 4,294,967,296$ of a non-zero fitness score (sampling with replacement). The expected fitness of a population that evolves using pure instinct is therefore approxi-mately 0. The values in Table 5 for pure learning come from Table 1 and the values for Experiment 1 come from Table 2. At this point, all we know is that a Baldwinian trajectory is superior to two other possible trajectories through bias space.

Table 5. A comparison of three trajectories through bias space.

|  | Expected Fitness after 10,000 Generations of a Population of 1000: | | |
| --- | --- | --- | --- |
| Noise Level | Pure Instinct $(\forall i)[s_i = 1]$ | Pure Learning $(\forall i)[s_i = 0]$ | Baldwin Baseline (Experiment 1) |
| 0.1% | 0.00 | 0.97 | 0.97 |
| 0.5% | 0.00 | 0.85 | 0.94 |
| 1% | 0.00 | 0.73 | 0.92 |

One advantage of our model (Section 4) over the model of Hinton and Nowlan (1987) is that we can manipulate bias strength independently of bias direction. In the following experiments, we examine what happens when (1) bias strength is fixed at 0.75, (2) bias strength is fixed at 0.5, (3) bias strength is fixed at 0.25, and (4) bias strength begins at 0, rises linearly to 1 at generation number 5,000, and then is held at 1 until generation 10,000. In these experiments, the genotype contained only the 32 bias direction genes; the 32 bias strength genes were removed from the genotype. Bias direction was allowed to evolve as usual. Our hypotheses were (1) of the three experiments with fixed bias, the experiment with bias fixed at 0.25 will have average fitness above 0.5 in the earliest generation, (2) of the four experiments, the experiment with bias ramped up linearly will have average fitness above 0.5 in the earliest generation, (3) the average fitness in the final generations will be greatest in the fourth case (bias ramped up linearly), and (4) of the three experiments with fixed bias, the experiment with bias fixed at 0.75 will have the greatest average fitness in the final generations. The first two hypotheses follow from the first aspect of the Baldwin effect and the second two hypotheses follow from the second aspect. Of the four trajectories, the fourth case (bias ramped up linearly) is the only one that is qualitatively in agreement with the prescriptions of the Baldwin effect. These hypotheses are supported in Figures 4, 5, 6, and 7, and in Table 6.

Table 6 shows that weaker bias accelerates evolution (columns 3 to 5), but stronger bias has the highest final fitness (column 6). Ramped bias (the last three rows) is the best of both worlds: it begins and ends with the highest fitness. These experiments support a *prescriptive* interpretation of the Baldwin effect.

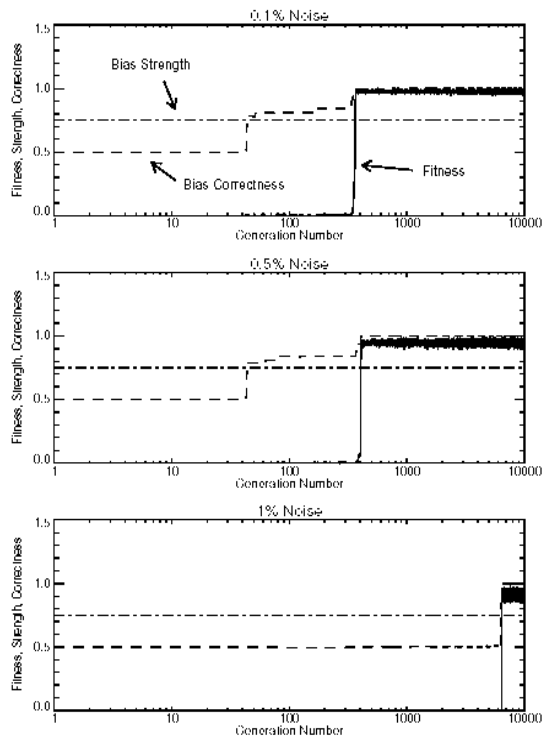

Figure 4. Experiment 3: Bias strength is fixed at 0.75.

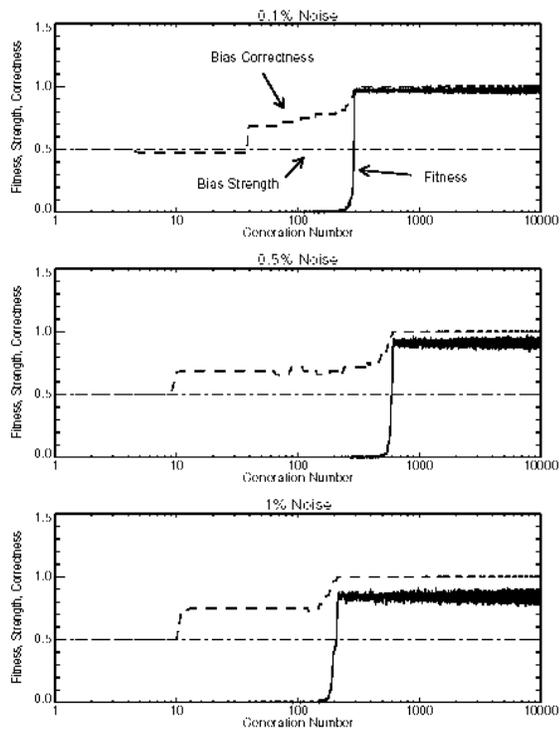

Figure 5. Experiment 4: Bias strength is fixed at 0.5.

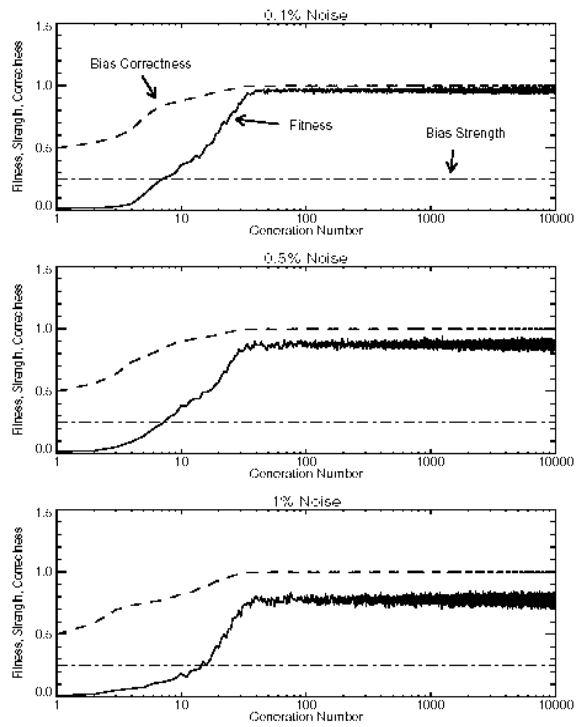

Figure 6. Experiment 5: Bias strength is fixed at 0.25.

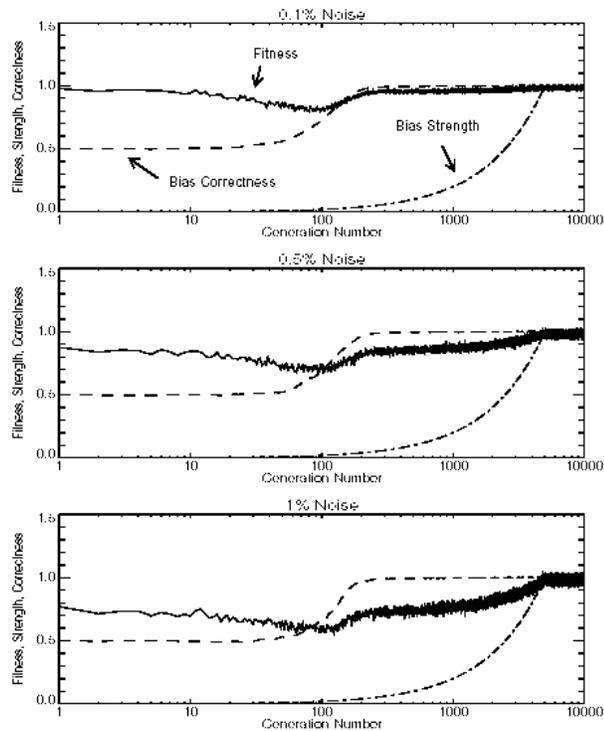

Figure 7. Experiment 6: Bias strength increases linearly from 0 in the first generation to 1 in generation 5,000. After generation 5,000, bias strength is held constant at 1.

Table 6. Rate of evolution and equilibrium state for forced bias strength trajectories.

| Bias Strength | Noise Level | First generation with: | | Number of generations difference | Average Fitness of the last 1000 generations |
| --- | --- | --- | --- | --- | --- |
| | | Fitness above 0.0 | Fitness above 0.5 | | |
| Fixed at 0.75 | 0.1% | 43 | 363 | 320 | 0.98 |
| $(\forall i)[s_i = 0.75]$ | 0.5% | 43 | 407 | 364 | 0.95 |
| | 1% | 6282 | 6443 | 161 | 0.91 |
| Fixed at 0.5 | 0.1% | 4 | 290 | 286 | 0.97 |
| $(\forall i)[s_i = 0.5]$ | 0.5% | 9 | 598 | 589 | 0.91 |
| | 1% | 10 | 208 | 198 | 0.84 |
| Fixed at 0.25 | 0.1% | 1 | 16 | 15 | 0.96 |
| $(\forall i)[s_i = 0.25]$ | 0.5% | 1 | 17 | 16 | 0.87 |
| | 1% | 1 | 24 | 23 | 0.77 |
| Ramped Linearly from 0 to 1 over 5,000 generations | 0.1% | 1 | 1 | 0 | 0.98 |
| | 0.5% | 1 | 1 | 0 | 0.98 |
| | 1% | 1 | 1 | 0 | 0.98 |

It is worth remarking on the fact that the final fitness with ramped bias is superior to the final fitness in Experiment 1 (Table 2). This is because the bias strength in Experiment 6 is not subject to genetic drift, since it is forced to be 1 in the last 5,000 generations. Only the bias direction is subject to genetic drift (which explains why the fitness is not 1.00). We would observe the same final fitness in Experiment 1 if we clamped the bias strength at 1 in the last 5,000 generations.

## 6. Discussion

Aside from their interest to students of the Baldwin effect, we believe that the above experiments are relevant for machine learning research in bias shifting algorithms and for evolutionary psychology.

## 6.1 How to Shift Bias

The machine learning community is moving away from the idea of a "universal learning algorithm", toward the recognition that all learning involves bias (Utgoff, 1986; Utgoff & Mitchell, 1982) and that the right bias must depend on the learner's environment (Schaffer, 1994; Wolpert, 1992, 1994). Recently several researchers have developed algorithms that shift their bias as they acquire more experience (Utgoff and Mitchell, 1982; Utgoff, 1986; Tcheng *et al.,* 1989; Schaffer, 1993; Gordon and desJardins, 1995; Bala *et al.*, 1995; Provost and Buchanan, 1995; Turney, 1995).

Research in shift of bias has not previously reported anything like the Baldwin effect. For example, Utgoff (1986) is mainly concerned with techniques for weakening a bias that is too strong, although there is a brief discussion of techniques for strengthening bias. Researchers have presented algorithms for searching bias space, but they have not studied the trajectories that their algorithms follow as they move through bias space. The Baldwin effect draws attention to the possibility of studying and understanding the trajectory of search through bias space. In addition to being descriptive (predictive), the Baldwin effect is prescriptive (normative). That is, the trajectory predicted by the Baldwin effect is not merely the *actual* trajectory of a bias shifting system (Section 5.2 and Section 5.3), it is also a *good* trajectory (Section 5.4). However, we have not demonstrated that it is an *optimal* trajectory.

Although our experiments use a genetic algorithm to search in bias space, the results do not seem to depend on the use of evolutionary computation. It seems likely that any robust, general-purpose search algorithm, such as tabu search (Glover, 1989, 1990) or simulated annealing (Lawrence, 1987), will follow a Baldwinian trajectory. An open issue is to discover the weakest possible (most general) conditions for the manifestation of the Baldwin effect. We believe that the lessons learned here apply generally to the research in machine learning in bias shifting algorithms, even when the algorithms do not use evolutionary computation. Supporting this hypothesis is an area for future work.

## 6.2 Evolutionary Psychology

It is becoming increasingly clear that human learning involves many instinctive elements (Barkow *et al.,* 1992; Pinker, 1994). Psychologists are moving away from a *tabula rasa* (blank slate) view of the mind, toward the view that learning is only possible on top of a substrate of instinct. We believe that learning and instinct are not two separate things; rather, they are opposite poles in a continuum. The model in this paper illustrates this idea. What we usually call *learning* might be better called *weakly biased learning*, and what we usually call *instinct* might be better called *strongly biased learning*. Although our model allows pure learning (no bias) and pure instinct (all bias), we believe that such extremes are not likely to occur, either in more complex animals (vertebrates) or in more complex machine learning applications.

The idea of varying degrees of bias strength may make it easier to understand how evolution can gradually move from instinct to learning or from learning to instinct. To replace a learned behavior with an instinctive behavior, evolution does not need to generate a "hard-wired" neural network that replaces a previous "soft-wired" neural network. Instead, evolution can gradually increase the bias strength in the network. (The details of how this can be done depend on the particular neural network model. See Section 3.4.)

# 7. Conclusion

This paper introduced a computational model that was devised to illustrate both aspects of the Baldwin effect and to make clear the implications of the Baldwin effect for the design of bias shifting algorithms. The model complements the work of Hinton and Nowlan (1987) and Harvey (1993). The genotype generalizes the Hinton and Nowlan (1987) genotype by separating bias strength from bias direction and allowing bias strength to be set to continuous values. The fitness function models the benefit of strong bias for reducing sensitivity to noise (Geman *et al.*, 1992), while Hinton and Nowlan's (1987) fitness function models the benefit of strong bias for reducing search.

Like Hinton and Nowlan's (1987) model, the model presented here is highly abstract and simple. It omits many elements of real learning systems, to make the interpretation of the experimental results easier. We have performed some preliminary experiments with a more complex bias shifting algorithm (Turney, 1995), which appears to display the same qualitative behavior as the simple model.

Hinton and Nowlan (1987) and Harvey (1993) assume a dichotomy: either there is a population of individuals that learn during their lifetime or there is a population of individuals that do not learn. In their model, learning is either on (**?**) or off (**1** or **0**). The lesson we can draw from their experiments is that it is better for individuals to learn.

In our model, bias strength can vary continuously. Evolution is a search through bias space and the search can follow many different kinds of trajectories. The Baldwin effect recommends a certain class of trajectories: those trajectories that begin with weak bias and gradually evolve towards strong bias. We can visualize the Baldwin effect as a wide band of paths through bias space. Hinton and Nowlan (1987) and Harvey (1993) restricted their attention to just two of the many paths through bias space (a Baldwinian path versus a no-learning path). The lesson we can draw from the experiments presented here is that some of the paths that wander outside of the band suggested by the Baldwin effect are inferior to some of the paths that stay inside the band (Section 5.4). We believe that thinking of the Baldwin effect in terms of a continuum of bias strength is richer and more fruitful than thinking in terms of a strict dichotomy of learning and instinct.

## Acknowledgments

Thanks to Darrell Whitley, Russell Anderson, Joel Martin, Martin Brooks, Diana Gordon, and three anonymous referees for their very helpful comments on earlier versions of this paper.

# Appendix: Machine Learning Bias Versus Statistical Bias

In this paper, we have used the definition of bias that is used in the machine learning research community (Utgoff and Mitchell, 1982; Utgoff, 1986; Haussler, 1988; Gordon and desJardins, 1995), which is slightly different from (but related to) the definition used by statisticians (Geman *et al.*, 1992). This appendix explains the relationships between these definitions.

In statistics, the bias of an estimator is the difference between the expected (mean) value of the estimator and the thing being estimated. In our model (Section 4), the $i$-th guess $g_i$ is an estimator for the $i$-th target $t_i$.

$$\text{statistical bias of } g_i = E(g_i) - t_i \qquad (27)$$

($E(\ldots)$ is the statistical expectation operator.) The statistical bias of $g_i$ depends on the relationship between the bias direction $d_i$ and the target $t_i$. Table 7 shows the relationship for each of the possible values of $d_i$ and $t_i$. We see that the behavior of the statistical bias depends on whether $d_i = t_i$. If we ignore the sign of the bias (take the absolute value), we have $(1 - s_i)p$ when $d_i = t_i$ and $(1 - s_i)p + s_i$ when $d_i \neq t_i$. Incorrect machine learning bias has the effect of adding an extra term $s_i$ to the magnitude of the statistical bias.

Table 7. The statistical and machine learning bias of the $i$-th guess.

| Bias Direction $d_i$ | Target Concept $t_i$ | Statistical Bias $E(g_i) - t_i$ | Machine Learning Bias Correctness | Machine Learning Bias Strength |
|---|---|---|---|---|
| 0 | 0 | $(1 - s_i)p$ | 1 | $s_i$ |
| 0 | 1 | $-(1 - s_i)p - s_i$ | 0 | $s_i$ |
| 1 | 0 | $(1 - s_i)p + s_i$ | 0 | $s_i$ |
| 1 | 1 | $-(1 - s_i)p$ | 1 | $s_i$ |

Statisticians are interested in the relationship between (statistical) bias and variance.

$$\text{var}(g_i) = E((E(g_i) - g_i)^2) \qquad (28)$$

Table 8 shows the variance of $g_i$ for each of the possible values of $d_i$ and $t_i$. Again, the behavior depends on whether $d_i = t_i$. Incorrect machine learning bias has the effect of adding an extra term $s_i(1-s_i)$ to the variance. However, the final term that is subtracted from the variance is a bit smaller when $d_i = t_i$:

$$2s_i p(1-p) \leq 2s_i p(2-p-s_i) \qquad (29)$$

This compensates somewhat for the extra term $s_i(1-s_i)$.

Table 8. The variance of the *i*-th guess.

| Bias Direction $d_i$ | Target Concept $t_i$ | Variance var($g_i$) |
|---|---|---|
| 0 | 0 | $p(1-p) + s_i p(1-s_i p) - 2s_i p(1-p)$ |
| 0 | 1 | $p(1-p) + s_i p(1-s_i p) + s_i(1-s_i) - 2s_i p(2-p-s_i)$ |
| 1 | 0 | $p(1-p) + s_i p(1-s_i p) + s_i(1-s_i) - 2s_i p(2-p-s_i)$ |
| 1 | 1 | $p(1-p) + s_i p(1-s_i p) - 2s_i p(1-p)$ |

When the machine learning bias strength is 1 ($s_i = 1$), the magnitude of the statistical bias is 0 if $d_i = t_i$ and 1 if $d_i \neq t_i$. The variance is 0, for both $d_i = t_i$ and $d_i \neq t_i$. That is, maximal machine learning bias strength results in minimal variance. Maximally strong and correct machine learning bias results in minimal statistical bias. Maximally strong and incorrect machine learning bias results in maximal statistical bias.

When the machine learning bias strength is 0 ($s_i = 0$), the magnitude of the statistical bias is $p$, for both $d_i = t_i$ and $d_i \neq t_i$. The variance is $p(1-p)$, for both $d_i = t_i$ and $d_i \neq t_i$. That is, minimal machine learning bias strength results in a low but non-minimal statistical bias (assuming $p$ is a small positive value, as it was in our experiments). It also results in a low but non-minimal variance.

The worst case for variance happens when the machine learning bias strength is 0.5. At $s_i = 0.5$, the extra term $s_i(1-s_i)$ reaches its maximum value. The variance is $\frac{1}{4}(2p-p^2)$ if $d_i = t_i$ and $\frac{1}{4}(1-p^2)$ if $d_i \neq t_i$. We have been assuming that $p < 0.5$. (If $p > 0.5$, then $\vec{\alpha}$ and $\vec{\beta}$ are more likely to be the negation of the target $\vec{t}$, so we would like the bias direction $\vec{d}$ to also be the negation of the target.) Therefore $2p < 1$, which means that $\frac{1}{4}(2p-p^2) < \frac{1}{4}(1-p^2)$. For small noise levels ($p$ near 0), $\frac{1}{4}(1-p^2)$ will be considerably larger than $\frac{1}{4}(2p-p^2)$. For example, if there is no noise, then the variance is $1/4$ if $d_i \neq t_i$ and 0 if $d_i = t_i$. This high variance may help to explain why there is selection pressure for weak (machine learning) bias when the (machine learning) bias correctness is low.

The reason that statisticians are interested in (statistical) bias and variance is that squared error is equal to the sum of squared (statistical) bias and variance.

$$\text{squared error} = E((g_i - t_i)^2) \tag{30}$$

$$E((g_i - t_i)^2) = (E(g_i) - t_i)^2 + E((E(g_i) - g_i)^2) \tag{31}$$

Therefore minimal (statistical) bias and minimal variance implies minimal squared error. (Note that the error $E((g_i - t_i)^2)$ is the error relative to the target $t_i$, not the error $E((g_i - \beta_i)^2)$ relative to the testing data $\beta_i$. The difference between these two errors is random noise, known as *irreducible prediction error*.) Typically there is a trade-off between bias and variance — decrease in one usually causes increase in the other — but we have seen here that it is sometimes possible to minimize both simultaneously.